\documentclass[10pt,twocolumn,letterpaper]{article}
\pdfoutput=1

\usepackage[pagenumbers]{cvpr} 

\usepackage{graphicx}
\usepackage{amsmath}
\usepackage{amssymb}
\usepackage{booktabs}

\usepackage[pagebackref,breaklinks,colorlinks]{hyperref}

\usepackage[capitalize]{cleveref}
\crefname{section}{Sec.}{Secs.}
\Crefname{section}{Section}{Sections}
\Crefname{table}{Table}{Tables}
\crefname{table}{Tab.}{Tabs.}

\usepackage{xcolor}

\def\cvprPaperID{9287} 
\def\confName{CVPR}
\def\confYear{2022}

\begin{document}

\title{MOSAIC: Mobile Segmentation via decoding Aggregated Information and encoded Context}


\author{Weijun Wang \qquad Andrew Howard\\
Google Research\\
{\tt\small weijunw, howarda@google.com}
}

\maketitle

\begin{abstract}
   We present a next-generation neural network architecture, MOSAIC, for efficient and accurate semantic image segmentation on mobile devices. MOSAIC is designed using commonly supported neural operations by diverse mobile hardware platforms for flexible deployment across various mobile platforms. With a simple asymmetric encoder-decoder structure which consists of an efficient multi-scale context encoder and a light-weight hybrid decoder to recover spatial details from aggregated information, MOSAIC achieves new state-of-the-art performance while balancing accuracy and computational cost. Deployed on top of a tailored feature extraction backbone based on a searched classification network, MOSAIC achieves a $5\%$ absolute accuracy gain surpassing the current industry standard MLPerf models and state-of-the-art architectures.
\end{abstract}

\section{Introduction}
\label{sec:intro}

Semantic image segmentation with the aim to predict semantic labels for every image pixel is one of the fundamental research topics in computer vision. With the remarkable development of deep convolutional neural networks in recent years, fully convolutional neural network (FCN) \cite{long2015} based semantic segmentation models \cite{chen2017,deeplabv3+,PSPnet} have significantly advanced the field compared to conventional approaches. Formulated as a dense prediction problem, semantic segmentation technique has a wide applications around autonomous driving, video surveillance, human-machine interaction, dense scene parsing and photo editing, etc.

To push the accuracy frontier, most state-of-the-art methods either choose to use deeper and larger backbone networks or introduce innovative operations in the encoder-decoder architecture with less concern about their computational cost or inference speed. One common practice is to adapt a strong backbone network built for image classification task to the segmentation task. To maintain the high-resolution feature representation for this dense prediction task, excessive down-sampling or striding operations at the late stage of the network are removed and the filter kernels are enlarged to retain the effective receptive field, which will inevitably bring larger computational costs. Another example exists in the symmetric encoder-decoder structure \cite{UNet,SegNet,DeconvNet}, many lateral skip connections, from the encoder network to the high-resolution feature representation at decoder counterpart, are added to recover the fine structure details. While more lateral skip connections show benefits in recovering the structural details, they also increase computation complexity and memory access cost. 

\begin{figure}[t]
\centering
\includegraphics[width=1.0\linewidth]{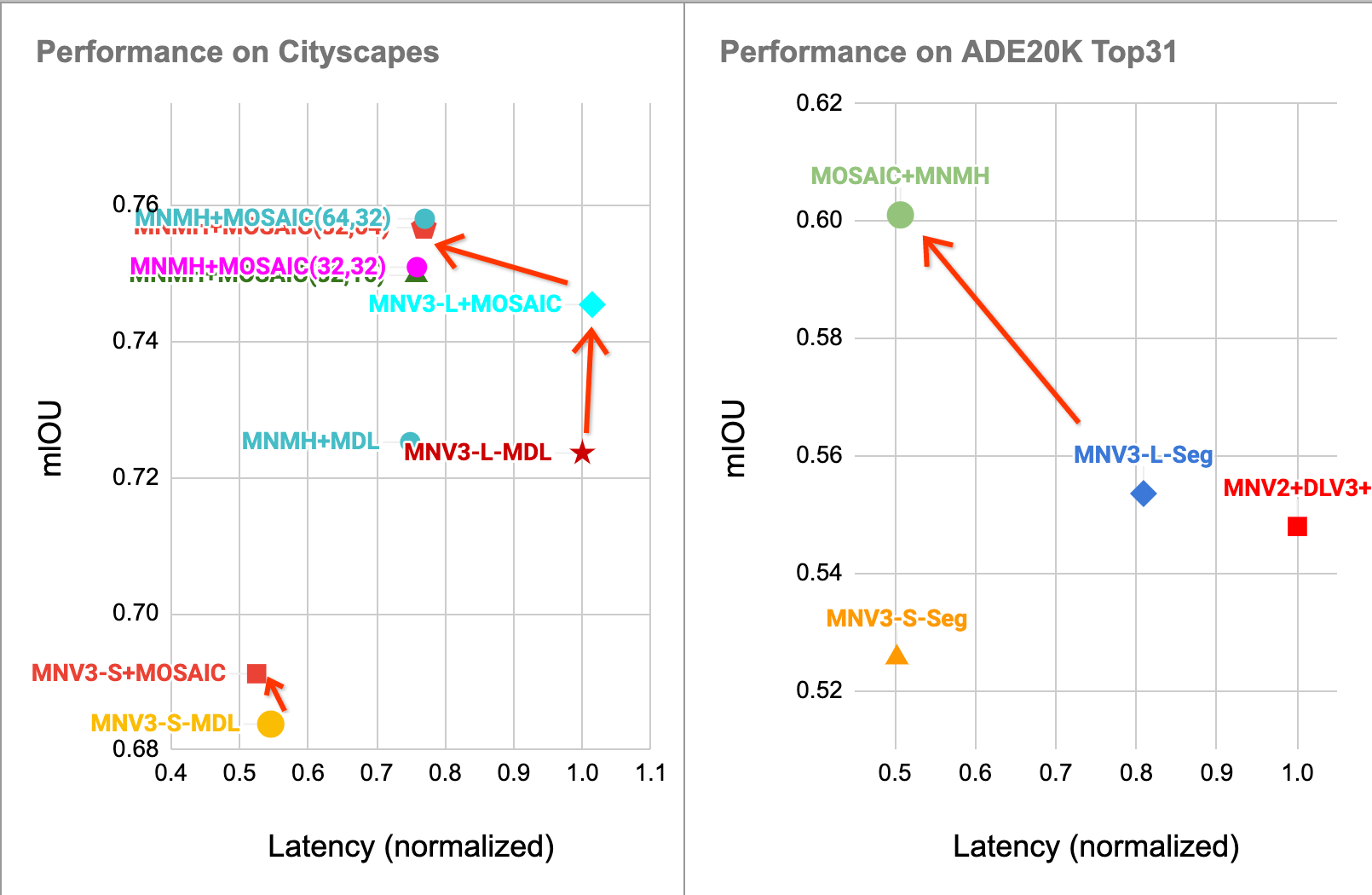}
\caption{Performance of MOSAIC on Cityscapes and ADE20K-Top31 in comparison with state-of-the-arts mobile segmenters. Latency is simply normalized using real on-device latency on various platforms as \cite{mobilenetmh}. Multiple MOSAIC models with different filter sizes are shown on the Cityscapes plot. Please see \cref{tab:cityscapes,tab:ade20k} for original data values.}
\label{fig:perf_plot}
\end{figure}

In recent years, to make the advanced segmentation models more applicable in real scenarios, real-time segmentation track has attracted more attention with a testing benchmark on desktop GPU, e.g. Nvidia Titan X. However, mobile platforms have strict constraints in computation resources and real-time algorithms on desktop GPU are not necessarily light-weighted or efficient enough for direct mobile deployment. Overall, there is still a shortage of mobile-friendly segmentation architectures considering the rising demand of efficient applications on mobile.

To accelerate a server-side segmentation model architecture for mobile deployment, common approaches include: (i) reducing input resolution, smaller input resolution can reduce computation cost at all stages of the network with the same network architecture; (ii) channel pruning, another straightforwards way to speedup inference, can be employed to reduce the number of feature channels both in the segmentation head architecture and  backbone network similar to \cite{mobilenetv1}. However, both acceleration approaches lead to a dramatic accuracy decrease, especially the performance degrades on fine structural details. In addition to the two common approaches above, MobileDeeplabv3 (R-ASPP) \cite{mobilenetv2} conduct segmentation head network pruning by removing atrous spatial pyramid pooling branches from the encoder head. To achieve a better trade-off between accuracy and efficiency for mobile applications, it is worth exploring the overall architecture design with a systematic approach balancing the functionality and the cost of the bottom-up encoding path and the top-down decoding path.

In this work, we propose an efficient neural network architecture for semantic image segmentation, which can be easily deployed on a wide variety of mobile platforms. Besides, mobile hardware platforms often use a diverse set of processors and accelerators (e.g. a mobile SoC containing CPU, GPU, DSP, NPU or EdgeTPU), each of which may have many unsupported neural operators. Therefore, we only consider commonly supported network operations by most mobile processor and accelerators in our design. With all constraints mentioned above, we try to re-balance the computational costs of different parts of the network architecture. A brief summary of MOSAIC's performance is shown in \cref{fig:perf_plot}. In the rest of the paper, we first briefly discuss related work in \cref{sec:related_work} and introduce MOSAIC's overall architecture in \cref{sec:encoder_decoder}. The design of a multi-scale context encoder is described in \cref{sec:context_encoder} and an efficient decoder module is described in \cref{sec:decoder}. Experimental results are provided in \cref{sec:exp} followed by a summary in \cref{sec:conclusion}.

\section{Related Work}
\label{sec:related_work}

\textbf{Semantic Image Segmentation.} Since the ground-breaking work \cite{long2015}, Fully Convolutional Neural Network (FCN) based methods have significantly improved state-of-the-art of semantic image segmentation. Building upon FCN by extending the idea of adding skip connections from low-level features to recover fine details, U-net \cite{UNet} and Segnet \cite{SegNet} proposed symmetric encoder-decoder architectures with feature map concatenation and pooling indice reusing respectively, which have achieved striking improvement on fine structures. By repeatedly applying a symmetric network with learned successive deconvolutions and unpooling on individual object proposals of various sizes and later combining them for final output, DeconvNet \cite{DeconvNet} is able to mediate the limitation of original FCN based methods in handling scale issues and fine details of small targets. However, the substantially increased computation in repeatedly inference with the network impedes its mobile applications.

\begin{figure*}
\vspace{-3mm}
    \centering
    \includegraphics[width=2.0\columnwidth]{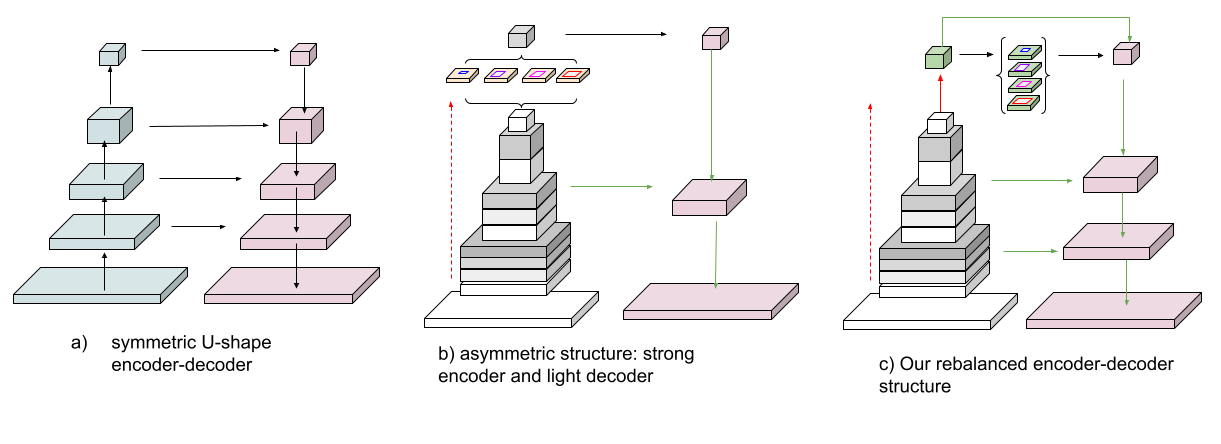}
    \vspace{-4mm}
    \caption{Various types of encoder-decoder structures for segmentation.}
    \label{fig:edt}
\vspace{-3mm}
\end{figure*}

Many state-of-the-art approaches have successfully adapted strong classification neural network architectures and classification pretraining techniques to pixel-wise dense prediction tasks. Such models usually come with an asymmetric encoder-decoder structure, emphasizing the feature encoding side inherited and pretrained from image classification. Their successes have demonstrated the importance of a strong and rich feature encoder. In particular, high-level semantics with rich contextual information are considered critical in recent success. To capture contextual information at various scales, Deeplabv3 \cite{chen2017} and its enhancement \cite{deeplabv3+} apply parallel atrous convolution with diverse rates, while PSPnet \cite{PSPnet} adapts spatial pyramid pooling \cite{SPP-DeepConvNet-he-eccv-14} in neural network for the dense scene parsing task.

\textbf{Towards Real-time Segmentation.} A wide spectrum of techniques have been proposed to speed up the FCN based segmentation models. As an early work for real-time segmentation, Enet \cite{Enet} designs an efficient architecture from the ground up using modified bottleneck residual blocks from \cite{ResNet}. With the idea of reducing computation by convolution factorization, ESPNet \cite{ESPNet} decomposes a standard convolution into spatial pyramid dilated convolutions following a pointwise convolution to enlarge the effective receptive field; the non-bottleneck residual block \cite{ResNet} is redesigned with 1D factorization to be used in a symmetric encoder-decoder network named ERFnet \cite{ERFNet}. ICNet \cite{ICNet} introduced an image cascade network with cascade feature fusion and cascade label guidance units to utilize semantic information in low resolution along with high-resolution image details to progressively and efficiently refine predictions. DLC \cite{DLC} employs a layer cascade network and performs regional convolutions only on hard regions feed-forward by earlier sub-models to reduce computation.
By applying a refinement idea similar to \cite{ladder} with focus on the encoding period, DFANet \cite{DFANet} exploits deep feature aggregation of sub-networks and sub-stages with multi-stage context.

\textbf{Efficient Network Architectures.} Network architecture design for better trade-off between accuracy and efficiency has been an active area in computer vision research. Group convolution \cite{alexnet} is utilized with channel shuffle operations in ShuffleNet \cite{ShuffleNet2017} and learned to keep useful dense connections in CondenseNet \cite{CondenseNet}. MobileNetV1 \cite{mobilenetv1} and Xception \cite{Chollet_2017_CVPR} explored depthwise separable convolution introducing a new trend in efficient networks; MobileNetV2 \cite{mobilenetv2} introduced a more efficient mobile network building block called inverted residual and linear bottleneck; MnasNet \cite{mnasnet} and MobileNetV3 \cite{mobilenetv3} explored neural architecture search to achieve improved accuracy and latency for mobile. Specially for mobile segmentation, a reduced form of Deeplabv3 named MobileDeeplabv3 is demonstrated in \cite{mobilenetv2}; a new efficient segmentation decoder Lite Reduced
Atrous Spatial Pyramid Pooling (LR-ASPP) is proposed in \cite{mobilenetv3}.

\section{Model Architecture Overview}
\label{sec:encoder_decoder}

FCN based semantic segmentation models \cite{long2015,SegNet,DeconvNet} usually consist of an encoder module and a decoder module. The encoder works
in a bottom-up process to convert low-level image features into high-level semantic features. To make the computation affordable, the encoder usually employs striding in convolution and pooling operations to reduce the spatial resolution while increasing the number of higher-level features. To the opposite, the decoder is usually designed to work in a top-down fashion which aims to recover as much spatial details as possible by considering encoded high-level semantic features in low-resolution.

The existing encoder-decoder architectures can be roughly divided into two types, see \cref{fig:edt} a) and b). The original encoder-decoder structure proposed by FCN has a symmetric U-shape, which balances the computation cost between the bottom-up encoding stage and the top-down decoding stage. In the symmetric structure, resolutions are gradually reduced in encoder while the spatial details are gradually recovered in decoder. Additionally, U-net \cite{UNet}, SegNet \cite{SegNet}, DeconvNet \cite{DeconvNet} use multiple lateral links to connect encoding layers with decoding layers at the same spatial resolution. The introduction of the lateral links combined with progressive decoding strategy is shown to be effective in improving quality especially on edges. At the meantime, the computational cost of such decoders is also relatively high, which is less feasible for mobile segmentation applications. In contrast, some other methods, e.g. Deeplabv3+ \cite{deeplabv3+}, put more focus on enhancing the encoder and uses a simple yet effective decoder as shown in \cref{fig:edt} b). One advantage of the asymmetric structure, compared to the symmetric one, is that many existing network backbones for classification and detection tasks can be easily adopted as the segmentation backbone.

To keep merits of both structures, we rebalance the computational cost of different parts and propose a more efficient asymmetric encoder-decoder structure with multiple lateral connections, shown in \cref{fig:edt} c). By using a more efficient encoder, we can shift some computation to the decoding stage to recover more details in a gradual manner. The detailed design of the encoder and decoder are described in \cref{sec:context_encoder,sec:decoder} respectively.

\section{Efficient Context Encoder}
\label{sec:context_encoder}

Contextual information is critical for many computer vision problems. In semantic segmentation, where pixel-wise dense prediction is performed, generating context features containing multi-scale information aligned with each pixel location is the key for encoder. Deeplab\cite{deeplabv3+} employs atrous spatial pyramid pooling (ASPP) using atrous convolutions with different
atrous rates. However, ASPP is computationally expensive for mobile applications. Besides, large atrous rates are not well supported on mobile accelerators.

In this work, multi-scale contextual information is explored and encoded using an efficient spatial feature pyramid module on top of efficient feature extractor backbone, \eg MobileNets. To further increase the diversity of context features, a multi-kernel group convolution module, containing parallel branches with different convolution kernel sizes, is applied after the spatial feature pyramid. Besides, efficient backbones for image classification can be modified with details in \cref{sec:backbone} for segmentation.


\begin{figure*}
\vspace{-3mm}
    \centering
    \includegraphics[width=2.0\columnwidth]{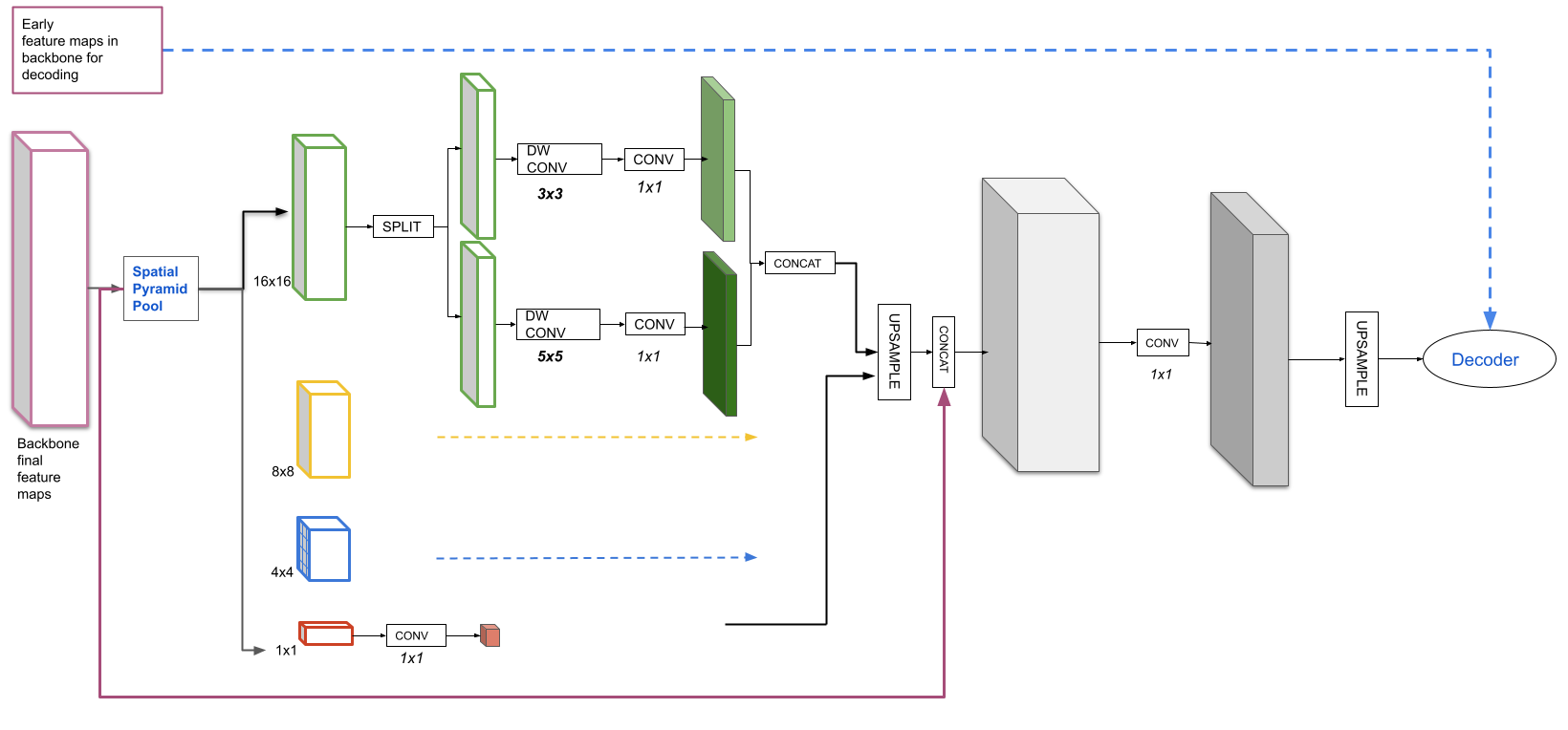}
    \caption{Context feature pyramid uses SPP and multi-kernel convolutions to efficiently encode pixel-wise context information.}
    \label{fig:context_encoder}
\vspace{-2mm}
\end{figure*}

\subsection{Contextual feature pyramid}
\label{sec:spp}

Feature pyramid network and spatial pyramid pooling (SPP) has been successfully applied as an information aggregation step for image classification, object detection \cite{SPP-DeepConvNet-he-eccv-14} thanks to its characteristic of generating a fixed-length output regardless of its input size. More importantly, a spatial feature pyramid (\eg \cite{FPN}) can capture contextual information at different scales and provide a diverse view-scopes, ranging from local to sub-regional and even a global reception field.

In this work, a spatial pyramid pooling (SPP) module is employed on top of a backbone network in order to collect contextual information from global levels to regional ones. The contextual feature maps produced by SPP have fixed spatial resolution regardless of the network input resolution. For instance, a 4-level pyramid consisting of 16$\times$16, 8$\times$8, 4$\times$4 spatial grids and a global pooling branch is shown in \cref{fig:context_encoder}. Each spatial bin in the pyramid provides a regional context for the pixels in its sub-region, while the global pooling branch generates representations of the whole scene.

\textbf{Multi-kernel group convolution}: To further increase the diversity of the contextual features, a \textit{multi-kernel group convolution} is proposed following the SPP step. The multi-kernel here refers to different convolution kernel sizes applied in parallel branches.
For example, 3$\times$3 and 5$\times$5 convolution kernels are used respectively in parallel shown in \cref{fig:context_encoder}. To reduce computation complexity, separable convolution, factoring a standard convolution into a depthwise convolution followed by a point-wise convolution (i.e. 1$\times$1 convolution), is used. In addition, we introduce group convolution into the multi-kernel convolution module and split feature channels into different groups to enforce channel independence among different groups and reduce computation. The primary purpose here is to use different kernel sizes in different groups in order to increase the diversity of contextual feature while keeping the computational cost down. An example of two feature groups are formed by equally splitting the input channels into two groups shown in \cref{fig:context_encoder}. At the end of the multi-kernel group convolution, feature maps from parallel branches are concatenated. Since the multi-scale contextual information is encoded in a fixed number of spatial pyramid bins regardless of the input resolution and the feature resolution has been greatly reduced during encoding, the additional multi-kernel group convolutions only take very limited computation.

\subsection{Pixel-wise context encoding}
\label{sec:cae}

In order to perform pixel-wise prediction, the multi-scale contextual features needs to be populated to each spatial location in the feature map for further aggregation. Using SPP, the relative spatial information is maintained during resolution reduction. To align features covering the same region, the standard bilinear interpolation is performed to upsample the reduced feature maps at each pyramid level to the original sizes before SPP. The upsampled contextual feature maps are then concatenated along the channel dimension. Besides, the original encoded features before feeding into the SPP step contain valuable information. In other words, the original encoded features from the backbone can be viewed as the bottom level of the contextual feature pyramid and are also concatenated. The concatenated feature maps are well aligned along the spatial dimension, while a diversity sets of contextual information are pooled together covering sub-regional, regional and global knowledge of the scene along the feature channels. To aggregate and further encode such diverse scene knowledge at each feature location, 1$\times$1 point-wise convolution is applied, which is shown on the right side of \cref{fig:context_encoder}. 

\section{Hybrid Decoder}
\label{sec:decoder}

The contextual knowledge output from the encoder has reduced resolution. To do dense pixel-level prediction at the original resolution, the original FCN \cite{long2015} uses a straightforward way is to directly up-sample the encoded context from a lower resolution to the original resolution. However, some detailed spatial information especially on edges, which has been lost during the striding convolutions and spatial pooling operations, cannot be recovered by such simple decoding method. Therefore, taking features with higher resolution from earlier stages of the backbone into decoding has shown to be effective in improving the segmentation quality  with additional computational cost.

\begin{figure}[t]
\centering
\includegraphics[width=1.0\linewidth]{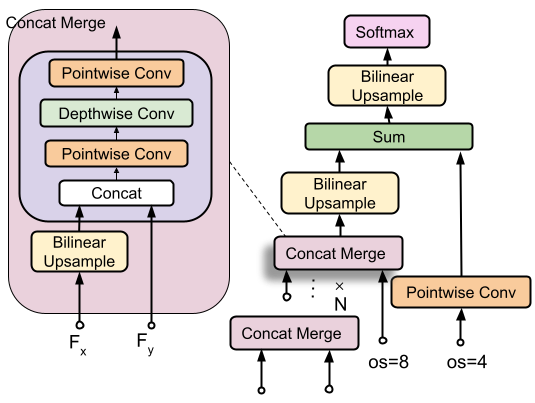}
\caption{A hybrid style decoder with concatenation merge and sum merge for efficient feature aggregation and fusion.}
\label{fig:fd}
\end{figure}

For mobile segmentation with a tight computation budget and limited availability of supported operations, any additional computation has to be used wisely. U-net \cite{UNet} style models employ a symmetric encoding and decoding structure and spend about half of the computation at the decoding stage. Although symmetric encoder-decoder structures have shown appealing results at edges, it is computationally prohibitive to directly apply in mobile use cases. In this work, we simply add a few skip connections from early feature layers and propose a lightweight decoder with a hybrid merging style. As an example, three skip routes are shown in \cref{fig:fd}. 

As shown in the overview architecture in \cref{fig:edt} c), although our overall model architecture is still asymmetric, more lateral skip connections are added to enhance the decoding process compared to the classic asymmetric encoder-decoder structure. We treat the decoding step as an iterative process between i) position-sensitive information aggregation and ii) semantic knowledge localization and propagation. The semantic knowledge localization and propagation is to spatially propagate the semantics stored in low-resolution feature maps precisely to higher-resolution feature maps, which can be carried out simply by bilinear upsampling without additional learnable parameters. Therefore, we keep the decoder blocks simple to make the training process easy to adapt with the fixed bilinear upsampling operations.

We introduce a simple hybrid decoder with two types of feature merging blocks: i) a concatenation merge block and ii) a summation merge block to decode from a rich set of aggregated context information. In practice, these two types of merge blocks can be applied in a mixed order.

In particular, the proposed concatenation merge module uses a standard concatenation step followed by a three-layer separable convolution block. In comparison to conventional concatenation merge modules(\eg \cite{deeplabv3+}), our separable convolution block, shown on the left of \cref{fig:fd}, consists of a serial of 1$\times$1 feature convolution, depthwise convolution and another 1$\times$1 feature convolution. Although this convolution block resembles the residual branch of blocks in \cite{mobilenetv2} and \cite{ResNet}, it is designed for a different purpose of feature decoding rather than bottleneck projection. The summation merge module operates with a standard element-wise addition, taking inputs from an upsampled semantic branch and a 1$\times$1 convolution on top of a spatial detail branch.

\section{Experimental Results}
\label{sec:exp}

\subsection{Tailored MobileNet backbone}
\label{sec:backbone}

As an advantage of our proposed asymmetric structure, classic network architectures designed for image classification can be easily adopted as the feature extraction backbones for segmentation task. Furthermore, we tailor the backbone networks to serve the segmentation task better similar to MobilenetV3-Deeplab segmentation\cite{mobilenetv3} with 3 modifications.

1) Excessive strides at late stages of a network are removed to lower the output stride. The output stride denotes the ratio of the input resolution to the output feature resolution. Specifically, we use final output stride of 16 for all backbones in all our training and evaluation. Unlike image classification task, larger output strides with excessive strides in convolution or pooling may hurt the accuracy of dense prediction in segmentation. 2) After removing excessive ending strides, dilated convolutions with dilation rate=2 at the ending stage is enabled to maintain the effective receptive field instead of enlarging kernels to keep computational cost low. 3) In addition, the number of feature channels is reduced by half and rounded to a multiplicative of 32 as mobile segmentation usually tackles a smaller number of classes and our experiments show its effectiveness. 

In particular, we made modifications on top of the most recent version of MobileNet, MobileNet Multi-Hardware (MNMH) in \cite{mobilenetmh}, which is searched for mobile classification with AutoML considering applications on various chips. The specifications are shown in \Cref{table:mn35a}.

\begin{table}[t]
\centering
\vspace{0pt}
\scalebox{0.85}{
\begin{tabular}{c|c|c|c|c|c|c}
\toprule[0.2em]
Input &           Operator & exp size& $\#out $ & $s$ & $os$\\
\toprule[0.2em]
$224^2\times3$  &  conv2d, $3\times3$   &  -   & 32    & 2   & 2\\
$112^2\times32$ &  bneck, $3\times3$    &  96  & 32    & 2   & -\\  
$56^2\times32$  &  bneck, $3\times3$    &  64  & 32    & 1   & 4\\   
$56^2\times32$  &  bneck, $5\times5$    &  160 & 64    & 2   & -\\  
$28^2\times64$  &  bneck, $3\times3$    &  192 & 64    & 1   & - \\  
$28^2\times64$  &  bneck, $3\times3$    &  128 & 64    & 1   & - \\ 
$28^2\times64$  &  bneck, $3\times3$    &  192 & 64    & 1   & 8 \\ 
$28^2\times64$  &  bneck, $5\times5$    &  384 & 128   & 2   & - \\ 
$14^2\times128$ &  bneck, $3\times3$    &  384 & 128   & 1   & -\\ 
$14^2\times128$ &  bneck, $3\times3$    &  384 & 128   & 1   & - \\ 
$14^2\times128$ &  bneck, $3\times3$    &  384 & 128   & 1   & - \\ 
$14^2\times128$ &  bneck, $3\times3$    &  768 & 160   & 1   & - \\ 
$14^2\times160$ &  bneck, $3\times3$    &  640 & 160   & 1   & - \\ 
$14^2\times160$ &  bneck, $3\times3$    &  960 & 192   & 1   & - \\ 
$14^2\times192$ &  bneck, $5\times5$    &  384 & 96    & 1   & - \\ 
$14^2\times96$  &  bneck, $5\times5$    &  384 & 96    & 1   & -  \\  
$14^2\times96$  &  bneck, $5\times5$    &  384 & 96    & 1   & - \\ 
$14^2\times96$  &  conv2d, $1\times1$   &  -   & $m$   & 1   & 16 \\ 
\midrule
$14^2\times m$  &  gpool, $14\times14$  &  -   & -     & 1   & - \\   
$1^2\times m $  &  conv2d, $1\times1$   &  -   & 1280  & 1   & - \\
$1^2\times 1280$&  conv2d, $1\times1$   &  -   & $k$     & 1   & - \\
\toprule[0.2em]
\end{tabular}
}
\caption{Specification for MobileNet Multi-Hardware after modification for segmentation task. The size of endpoint filters is denoted as $m$, 3 values $(448, 480, 512)$ of which are used in our experiments. $s$ denotes convolution stride. $os$ denotes output stride, the ratio of the input resolution to the output feature resolution. $bneck$ denotes MobileNetV2's inverted bottleneck block; $exp$ $size$ means the output filter size of the expansion layer. ReLU is used as the activation function in all convolution layers, where batch normalization is enabled and bias is not used. The last 3 rows are not used for segmentation task but for pretraining on ImageNet classification. Input resolution can be set as needed, which is set as $224\times224$ for illustration purpose.}
\label{table:mn35a}
\end{table}

\subsection{Implementation Details}
\label{sec:implementation_details}

\textbf{Encoder-decoder details}: Regarding the default parameter choices of the encoder architecture, we use a 3-level spatial pyramid pooling consisting of 16$\times$16, 8$\times$8 and 4$\times$4 bins respectively without a global pooling branch; in the multi-kernel group convolution, two groups with equal splitting are used followed by separable convolutions with kernel sizes of 3$\times$3 and 5$\times$5 respectively. For the decoder architecture, two additional skip connections are employed as shown in \cref{fig:edt} besides the skip connection in encoder. The skip link from output stride 8 of the backbone connects the decoder with a concatenation merge, while the final skip link from output stride of 4 use a summation merge to recover the final structure details. In following experiments, we explored different structure settings by changing one parameter at a time and keep the rest as default.

\textbf{Latency measurement} is conducted with TFLite models on device with the original image resolution of each dataset. In particular, we pick Pixel 4, one mid-tier mobile device which uses CPU, GPU, DSP and EdgeTPU. For each latency measurement, 10 separate on-device benchmark runs are conducted and the average result is reported to reduce noise in bench-marking.

\begin{table}[t]
  \centering
  \scalebox{0.75}{ 
  \begin{tabular}{l | c | c | c }
    \toprule
    Mobile Segmenter & mIOU (\%) & MAdds (B) & Latency (ms)  \\
    \midrule
    \midrule
    SegNet \cite{SegNet} & 57.0 & 82 & - \\
    ICNet \cite{ICNet} & 69.5 & 31 & - \\
    ERFNet \cite{ERFNet} & 69.7 & 26 & - \\
    ESPNet \cite{ESPNet} & 60.3 & 4.5 & - \\
    ENet \cite{Enet} & 58.3 & 3.8 & - \\
    \midrule
    \midrule
    MNV3-small + MDL\cite{mobilenetv3} & 68.38 & 2.90 & 476 (CPU) \\ 
    &  &  &  379 (GPU) \\
    &  &  &  291 (DSP) \\ 
    \midrule
    MNV3-small + MOSAIC & 69.12 & \textbf{3.02} & 518 (CPU)  \\
     &  &  & 323 (GPU) \\
     &  &  & 317 (DSP) \\
    \midrule
    \midrule
    MNV3-large + MDL\cite{mobilenetv3} & 72.36 & 9.74 & 1317 (CPU) \\ 
     &  &  &  430 (GPU) \\
     &  &  &  736 (DSP) \\
    \midrule
    MNV3-large + MOSAIC & 74.54 & 9.83 & 1192 (CPU) \\
     &  &  & 427 (GPU) \\ 
     &  &  & 843 (DSP) \\
    \midrule
    \midrule
    MNMH + MDL\cite{mobilenetv3} & 72.52 & 20.71 & 899 (CPU)\\ 
     &  &  & 546 (GPU) \\ 
     &  &  & 217 (DSP) \\ 
    \midrule
    MNMH + MOSAIC & \textbf{75.67} & 20.86 & 881 (CPU)\\
     &  &  & 564 (GPU) \\
     &  &  & 241 (DSP) \\
    \midrule
    \bottomrule
  \end{tabular}
  }
  \caption{Semantic segmentation results on Cityscapes val set. We use MNMH with $m$=480 and set (32, 64) for \#filters in MOSAIC's encoder and decoder respectively. MNV3 denotes MobileNetV3 and MDL denotes Mobile-Deeplab in \cite{mobilenetv3}. \cref{fig:perf_plot} Cityscapes is based on numbers in this table. MAdds and latency are measured with original input resolution of 1024x2048.}
  \label{tab:cityscapes}
\end{table}

\begin{table}[t]
  \centering
  \scalebox{0.85}{
  \begin{tabular}{c | c| c | c }
    \toprule
    SPP encoder & decoder  & mIOU(\%) & MAdds(B)  \\
    \#filter    & \#filter &          &           \\
    \midrule
    \midrule
    32 & 16 & 74.97 & 20.31 \\
    32 & \textcolor{green}{32} & \textcolor{green}{75.09} & 20.60 \\
    32 & \textcolor{green}{64} & \textbf{\textcolor{green}{75.67}} & 20.86 \\
    \midrule
    64 & 16 & 74.76 & 21.01 \\
    64 & \textcolor{green}{32} & \textcolor{green}{75.80} & 21.19 \\
    64 & \textcolor{pink}{64} & \textcolor{pink}{75.15} & 21.55 \\
    64 & \textcolor{pink}{128} & \textcolor{pink}{74.65} & 22.29 \\
    \midrule
    \midrule
    16 & 64 & 75.04 & 20.59 \\
    \textcolor{green}{32} & 64 & \textcolor{green}{75.67} & 20.86 \\
    \textcolor{pink}{64} & 64 & \textcolor{pink}{75.15} & 21.55 \\
    \textcolor{pink}{128} & 64 & \textcolor{pink}{74.58} & 23.55 \\
    \bottomrule
  \end{tabular}
  }
  \caption{Results on Cityscapes by adjusting filter sizes in encoder-decoder of MOSAIC head. We use MNMH with $m$=480 as the backbone. With the same \#filters (e.g. 32 or 64) in SPP encoder, enlarging \#filters in decoder from 16 to 32 (or even 64) \textcolor{green}{improves} the accuracy; however, excessive enlargement of \#filters in decoder (e.g. to 128 or 64) results in \textcolor{pink}{drop} in accuracy. Similarly, accuracy is increased by enlarging \#filters from 16 to 32 in the SPP encoder with the same decoder \#filters (e.g. 64); however, increasing \#filters in SPP encoder from 32 to 128 (or sometimes just 64) becomes excessive and leads to \textcolor{pink}{drop} in accuracy, comparing between lines with the same decoder \#filters.}
  \label{tab:filters_cityscapes}
\end{table}

\subsection{Results on Cityscapes}

The Cityscapes dataset \cite{Cordts2016Cityscapes} is a widely used segmentation dataset with an outdoor city street setting, consisting of 19 foreground classes and a background class. It contains pixel-level high quality fine annotations of 5000 images (2,975 in train, 500 in validation and 1,525 in test set) and additional 20000 coarsely annotated images. The original image resolution is 1024$\times$2048. We only exploit the fine annotations and use mIOU (the mean Intersection-Over-Union) as the accuracy metric. 

\begin{table}
  \centering
  \scalebox{0.85}{
  \begin{tabular}{c | l | c | c | c }
    \toprule
    \#Pyramid Levels & \#Bins & GC & mIOU (\%) & MAdds (B) \\
    \midrule
      & [1,4] & Y & 74.67 & 20.79 \\
    2 & [4,8] & Y & 75.30 & 20.81 \\
      & [4,16] & Y & 74.65 & 20.81 \\
      & [8,16] & Y & 73.52 & 20.82 \\
    \midrule
    3 & [4,8,16] & Y & \textbf{75.67} & 20.86 \\
      &          & N & 74.29 & 20.87 \\
    \midrule
    4 & [1,4,8,16] & Y & 75.04 & 20.88 \\
    \bottomrule
  \end{tabular}
  }
  \caption{Results of adjusting the spatial pyramid pooling structure in MOSAIC encoder on Cityscapes. GC denotes using group convolution. When group convolution is enabled, we only use 2 groups. Group convolution is shown to be effective to improve accuracy and also reduce computation. The best accuracy is achieved with a 3-level pyramid.}
  \label{tab:spp_cityscapes}
\end{table}

\begin{table}[t]
  \centering
  \scalebox{0.85}{
  \begin{tabular}{c | c | c | c }
    \toprule
    \#Skip connections & Merge styles & mIOU (\%) & MAdds (B) \\
    \midrule
    0 & -   & 72.62 & 20.108 \\
    \midrule
    1 & 4-S & 73.57 & 20.270\\
      & 4-C & 74.19 & 21.465\\
      & 8-C & 74.51 & 20.708\\
    \midrule
    2 & 8-C, 4-S & \textbf{75.67} & 20.860\\
      & 8-S, 4-S & 73.79 & 20.390\\
      & 8-C, 4-C & 75.19 & 21.685\\
    \midrule
    3 & 8-C,4-S,2-S & 75.35 & 21.186\\
    \bottomrule
  \end{tabular}
  }
  \caption{Results of using different decoder structure settings in MOSAIC on Cityscapes. 4-S denotes sum merge with $os$=4, 8-C denotes concatenation merge with $os$=8.}
  \label{tab:decoder_cityscapes}
\end{table}

\begin{table}
  \centering
  \scalebox{0.85}{
  \begin{tabular}{l | c | c | c }
    \toprule
    Backbone & head & mIOU (\%) & MAdds (B) \\
    \midrule
    MNV3-small &                 & 69.12 & 3.02  \\ 
    MNV3-large &                 & 74.54 &  9.83 \\
    MNMH ($m=448$) & MOSAIC  & 73.96 & 20.81 \\ 
    MNMH ($m=480$) &         & \textbf{75.67} & 20.86 \\
    MNMH ($m=512$) &         & 75.22 & 20.90 \\ 
    \midrule
    \midrule
    ResNet50   & MOSAIC & 76.34 & 302.79 \\
    ResNet101  &        & 76.66 & 458.13\\
    \midrule
    Xception71 & MOSAIC & 79.71 & 466.46\\
               & DLV3+\cite{deeplabv3+}  & 79.55 & 628.81 \\ 
    \bottomrule
  \end{tabular}
  }
  \caption{Results of using different backbone feature extractors with MOSAIC on Cityscapes val set. Xception71 is a modified Xception variant described in DeepLabV3+\cite{deeplabv3+} denoted by DLV3+. MOSAIC outperforms \cite{deeplabv3+} using the same Xception71 backbone.}
  \label{tab:backbones_cityscapes}
\end{table}

All models are evaluated with a single-scale input because inference strategies, including multi-scale inputs and additional left-right flipped images will increase the computation cost significantly and are not applicable in on-device mobile use cases. For training, we first pretrain the tailored backbone with an additional global pooling layer and two fully connected layers in the end (which are removed for segmentation, see \cref{table:mn35a}) on ImageNet dataset until converge. Then we follow the same end-to-end segmentation training protocol as \cite{deeplabv3+}. In short, we employ the "poly" learning rate decay schedule and data augmentation in random scale and flip during training. We conducted a sweep search for initial learning rate in the range between 5e-4 and 2e-3 and report the best accuracy for each specification. Specially, we use training crop size 769$\times$769 with batch size of 8 to fit in P100 GPU. The results on Cityscapes validation set are reported in \cref{tab:cityscapes,tab:filters_cityscapes,tab:spp_cityscapes,tab:decoder_cityscapes,tab:backbones_cityscapes}.

\subsubsection{Performance Analysis on Cityscapes}

We explored how different settings of architecture affect model performance on Cityscapes dataset, especially those in MOSAIC's head architecture. In \cref{tab:filters_cityscapes}, we studied performance changes with respect to changes of \#filters in encoder and decoder. The results show that the differences in mIOU accuracy are within $2\%$ and changes in MAdds are also small even with 4x differences in \#filters. This demonstrates that MOSAIC is a lightweight segmentation head architecture and only takes a small portion of total computation compared to the backbone. In \cref{tab:spp_cityscapes}, we explored various settings of the spatial pyramid pooling structure in MOSAIC's context encoder, including the number of pyramid levels and pooling bins. In particular, a 3-level pyramid with 4$\times$4,8$\times$8 and 16$\times$16 bins performances the best. Besides, group convolution in the encoder, which reduces the computation, is found to be effective in improving accuracy. Interestingly, accuracy drops by adding an additional level with global pooling to the 3-level pyramid. In \cref{tab:decoder_cityscapes}, we explored decoder structure with various numbers of skip connections and feature merge styles. We found that a decoder with 2 skip connections using a concat feature merge at $os$=8 and a sum feature merge at $os$=4 gives a relatively good result. In \cref{tab:backbones_cityscapes}, we summarized MOSAIC's performance with various backbones.

\subsection{Results on ADE20K}

ADE20K \cite{ade20k} is another popular segmentation and scene parsing dataset covering stuff, objects and object parts, which has been adopted as a benchmark dataset for mobile segmentation by MLPerf v1.0 \footnote{\url{https://https://mlcommons.org/en/inference-mobile-10/}}. We follow the same protocol of MLPerf v1.0 by training and evaluating with the top-31 classes instead of the original 150 classes. A single-scale input with resolution of 512$\times$512 is used in our evaluation. Our results are compared with other top mobile segmenters including the MLPerf v1.0 standard model in \cref{tab:ade20k}, showing a substantial improvement and better trade-offs between accuracy and latency. Notably, MOSAIC achieves 5\% absolute gain in mIOU while keeping the on-device latency low especially on Pixel4 DSP and EdgeTPU.

\begin{table}
  \centering
  \scalebox{0.75}{ 
  \begin{tabular}{l | c | c | l}
    \toprule
    Mobile Segmenter & mIOU (\%) & MAdds (B) & Latency(ms)  \\
    \midrule
    MNV2+DLV3+\cite{deeplabv3+} & 54.80 & 2.7 & 219 (CPU)  \\
    (MLPerf v1.0)  &  &  & 134 (GPU) \\
      &  &  & 109 (DSP)\\ 
      &  &  & 165.6 (EdgeTPU) \\
    \midrule
    MNV3-small + LR-ASPP\cite{mobilenetv3} & 52.57 & 0.6 & 99.7 (CPU) \\ 
      &  &  & 67.08 (GPU) \\
      &  &  & 71.34 (DSP)\\
      &  &  & 66.23 (EdgeTPU) \\
    \midrule
    MNV3-large + LR-ASPP\cite{mobilenetv3} & 55.36 & 1.43 & 176.1 (CPU) \\  
      &  &  & 63.72 (GPU) \\
      &  &  & 135 (DSP) \\
      &  &  & 118.8 (EdgeTPU) \\
    \midrule
    MOSAIC + MNMH & \textbf{60.10} & 2.98 & 116 (CPU) \\ 
    (ours)  &  &  & 107 (GPU) \\
      &  &  & 39 (DSP) \\
      &  &  & 56.6(EdgeTPU) \\
    \bottomrule
  \end{tabular}
  }
  \caption{Semantic segmentation results on ADE20K-Top31 val set. MNV2 denotes MobileNetV2 \cite{mobilenetv2}. DLV3+ denotes DeepLabV3+ \cite{deeplabv3+}. MNV3s denotes MobileNetV3-small. MNV3-large denotes MobileNetV3-large. We employed a tailored segmentation backbone MNMH described in \cref{sec:backbone} with $m$=448. \#filter in SPP encoder and in decoder are both set as 64. Our model achieves 5\% absolute gain in mIOU while keeping the on device latency low especially on DSP and EdgeTPU of Pixel 4.}
  \label{tab:ade20k}
\end{table}

\begin{table}[t]
  \centering
  \scalebox{0.82}{ 
  \begin{tabular}{c | c| c } 
    \toprule
    SPP encoder & decoder & mIOU (\%) \\ 
    \#filter & \#filter & \\ 
    \midrule
    \midrule
    16 & 16 & 58.59   \\ 
    16 & \textcolor{green}{32} & \textcolor{green}{59.53} \\ 
    16 & \textcolor{green}{64} & \textcolor{green}{59.92} \\ 
    \midrule
    32 & 16 & 58.78  \\ 
    32 & \textcolor{green}{32} & \textcolor{green}{59.64} \\ 
    32 & \textcolor{green}{64} & \textcolor{green}{59.86}  \\ 
    32 & \textcolor{pink}{128} & \textcolor{pink}{59.23}  \\ 
    \midrule
    64 & 16 & 58.83 \\ 
    64 & \textcolor{green}{32} & \textcolor{green}{59.62}  \\ 
    64 & \textcolor{green}{64} & \textbf{\textcolor{green}{60.10}} \\
    64 & \textcolor{pink}{128} & \textcolor{pink}{59.66}  \\ 
    \bottomrule
  \end{tabular}
  }
  \caption{Results on ADE20K-Top31 by adjusting filters in the encoder-decoder head of MOSAIC. MNMH($m$=448) is used as the feature extractor backbone. With the same \#filters in decoder, increasing \#filters in SPP encoder from 16 to 32 (or even 64) achieves better (or at least similar) accuracy (to read by comparison between blocks). Similarly, with the same \#filters in SPP encoder, increasing \#filters in decoder from 16 to 32 (or even 64) results in consistent \textcolor{green}{gain} in accuracy; however, further doubling \#filters in decoder from 64 to 128 leads to \textcolor{pink}{drop} in accuracy.}
  \label{tab:filters_ade20k32}
\end{table}

We also studied the effects in model performance by adjusting encoder-decoder filters in MOSAIC head, summarized in \cref{tab:filters_ade20k32}. Overall, we observed small and smooth changes in performance with different filter settings. Detailed observations are provided under \cref{tab:filters_ade20k32}.

\section{Conclusion and future work}
\label{sec:conclusion}

We presented a next-generation neural network architecture, MOSAIC, for efficient and accurate semantic image segmentation, which is deployable across different mobile chips and hardware platforms. We introduced a simple yet effective asymmetric encoder-decoder structure with an efficient multi-scale context encoder and a light-weight hybrid decoder. Our extensive experiments have shown that MOSAIC achieves new state-of-the-art performance with balanced accuracy and computational cost. In future work, we plan to explore AutoML to further improve segmentation architecture.

\section{Acknowledgement}

 This project would not have been possible without the support from Hartwig Adam and Google Mobile Vision team, Google Research Perception and Cerebra. We would like to thank Jaeyoun Kim, Liang-Chieh Chen, Alex Stark, Grace Chu and Yukun Zhu for helpful discussion. Many thanks to the feedback from MLCommons MLPerf mobile working group members including David Kanter, Google Silicon team, MediaTek (Koan-Sin T an, Code Lin, Yu Tseng, and Cheng-Ming Chiang), Qualcomm Technologies Inc.(William Chou, Erich Plondke, and Mark Charlebois), and Samsung System LSI (Jungwook Hong, Mostafa El-Khamy, Jaegon Kim and Hai Su).

{\small
\bibliographystyle{ieee_fullname}
\bibliography{egbib}
}

\end{document}